%% file: main.tex
\documentclass[runningheads]{llncs}

\usepackage{eccv}

\usepackage{eccvabbrv}

\usepackage{graphicx}
\usepackage{booktabs}
\usepackage{float}
\usepackage{adjustbox}

\usepackage{placeins}

\usepackage[accsupp]{axessibility}  

\usepackage{hyperref}

\usepackage{orcidlink}

\usepackage{siunitx}
\usepackage{xcolor}
\usepackage{pifont}
\usepackage{multirow}

\newcommand{\cmark}{\ding{51}}

\AtEndPreamble{%
  \crefname{figure}{Fig.}{Figs.}%
  \Crefname{figure}{Figure}{Figures}%
  \crefname{equation}{Eq.}{Eqs.}%
  \Crefname{equation}{Equation}{Equations}%
  \crefname{appendix}{Appendix}{Appendices}%
  \Crefname{appendix}{Appendix}{Appendices}%
}
\newcommand{\dataset}{GEOID-Flood }
\newcommand{\datasetns}{GEOID-Flood}

\begin{document}
\title{\datasetns: A Large-Scale Multi-Modal Benchmark Dataset for Flood Segmentation}

\titlerunning{GEOID-Flood: Multi-Modal Flood Segmentation Benchmark}

\author{Gaetano Chiriaco\inst{1}\orcidlink{0009-0004-8580-8806} \and Luca Barco\inst{1,2}\orcidlink{0000-0002-9089-9616} \and Andrea Bragagnolo\inst{1}\orcidlink{0000-0002-8619-1586} \and Claudio Rossi\inst{1}\orcidlink{0000-0001-5038-3597} \and Edoardo Arnaudo\inst{1}\orcidlink{0000-0001-9972-599X}}

\authorrunning{G.~Chiriaco et al.}

\institute{Fondazione LINKS, Via Pier Carlo Boggio, 61, 10138 Torino, Italy \\
\email{\{name\}.\{surname\}@linksfoundation.com} \\
\and
Politecnico di Torino, Corso Duca degli Abruzzi 24, 10129 Torino, Italy \email{\{name\}.\{surname\}@polito.it} 
}

\maketitle

\input{sections/0_abstract}
\input{sections/1_intro}
\input{sections/2_relworks}
\input{sections/3_dataset}
\input{sections/4_method}
\input{sections/5_results}
\input{sections/6_concl}

\section*{Acknowledgements}
This study was carried out in the context of the SIU (CUP I53D24000060005) and REHUBS (grant number 101214051) projects.

%
%
\bibliographystyle{splncs04}
\bibliography{main}

\clearpage
\appendix
\input{sections/X_suppl}

\end{document}

%% file: sections/0_abstract.tex
\begin{abstract}
Geospatial foundation models aim to learn representations that transfer across regions and sensors, yet evaluating them on specific tasks requires large, high-quality, multi-modal benchmarks that measure how well such models extract value from data.
Concerning flood mapping, existing datasets rarely combine bi-temporal SAR and co-registered optical imagery at scale, leaving the value of foundation models for this downstream task largely untested.
We introduce \datasetns\footnote{\url{https://github.com/links-ads/geoid-flood}}, a large-scale multi-modal flood segmentation benchmark, derived from Copernicus Emergency Management Service activations, spanning 219 events across 65 countries over ten years.
The dataset provides more than 14\,000 tiles with co-registered pre- and post-event Sentinel-1, in GRD and RTC format, pre-event Sentinel-2 composite, and DEM, including manually validated labels that separate background from permanent water and flooded water.
Using this benchmark, we evaluate foundation models against conventional encoders across single-image, multi-temporal, and multi-modal protocols. We report three main findings: foundation models offer a consistent but modest advantage; optical–SAR fusion with finetuning best resolves transient flooding; and models trained on \dataset{} transfer to unseen events better than those trained on existing datasets.
\keywords{Flood Segmentation \and Synthetic Aperture Radar \and Geospatial Foundation Models \and Benchmark Dataset}
\end{abstract}

%% file: sections/1_intro.tex
\section{Introduction}
\label{sec:intro}

Floods are among the most frequent and damaging natural hazards, affecting more
people worldwide than any other weather-related disaster, and both their
frequency and severity are projected to rise under a warming
climate~\cite{cred2020,ipcc2021}.
To reduce and quantify their impact, there is a growing need to map flood extent rapidly and over wide areas.
Remote Sensing has become the backbone of operational
flood monitoring~\cite{martinis2009,twele2016}, where mapping must stay reliable
under cloud cover, at night, and across wide geographic extents.
Synthetic Aperture Radar (SAR) meets these requirements, but open water is
intrinsically ambiguous in Sentinel-1 backscatter: a single acquisition rarely
separates permanent rivers and reservoirs from newly inundated terrain.
Resolving this ambiguity demands temporal context (pre/post-event change), together with
annotations that distinguish transient flooding from both background soil and
permanent water bodies.
Meeting these requirements at operational scale, i.e., across diverse regions, events, and sensors, points to two needs: representations that generalize beyond their training conditions, and benchmarks rich enough to evaluate whether they do.



Deep learning, and in particular geospatial foundation models pretrained on large Earth-observation corpora, offers a promising answer to the first. The second, however, remains under-explored: existing datasets rarely meet the criteria needed to evaluate such models \cite{doerksen2026earthshift}. To our knowledge, no public dataset combines bi-temporal SAR for pre- and post-event scenes, co-registered optical imagery for multi-modal fusion, and event-level train/test splits at continental scale.

\begin{figure*}[t]
  \centering
  \includegraphics[width=\linewidth]{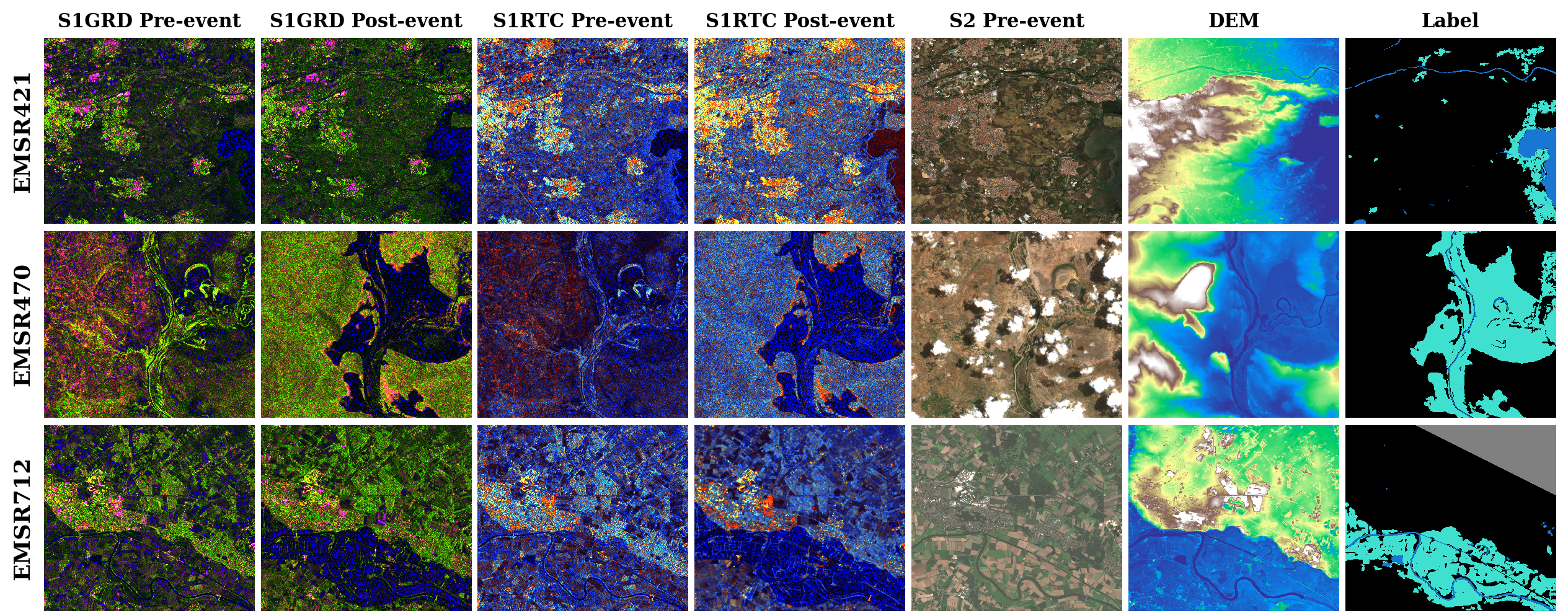}
  \caption{Representative tiles from three \dataset events. From left to right: pre- and post-event Sentinel-1 GRD and RTC (VV/VH format), pre-event Sentinel-2 (RGB), DEM, and label. Flooded water is shown in cyan, permanent water in blue, invalid pixels in gray.}
  \label{fig:dataset_overview}
\end{figure*}

We address this gap with \dataset(Geospatial Earth Observation Imagery Dataset for Floods), a large-scale flood benchmark derived from Copernicus Emergency Management Service (CEMS) activations. It covers \SI{1141749}{\kilo\metre\squared} of flood-affected terrain, nearly double the largest prior dataset and the widest spatial extent reported to date over the longest acquisition window so far (2016--2026). Unlike existing benchmarks, which typically provide one or two sensors, \dataset jointly offers bi-temporal Sentinel-1 (S1), in GRD and RTC variants, Sentinel-2 (S2), and a Digital Elevation Model (DEM), and pairs these inputs with a dedicated permanent water layer, so that transient flooding is annotated separately from background and permanent water (\cref{fig:dataset_overview}).

This design lets us study representation quality and temporal modelling in a single, controlled setting. We organize our analysis around four research questions that link the dataset (\cref{sec:dataset}) to the methodological choices (\cref{sec:method}) and experiments (\cref{sec:results}):

\begin{itemize}
    \item \textbf{RQ1 (pretraining):} \textit{how do geospatial foundation models compare to other pretrained encoders, and does finetuning beat frozen features?}

    \item \textbf{RQ2 (temporal):} \textit{Is modelling the flood class with a specific loss or architecture better than deriving it with post-hoc water-body delineation?}

    \item \textbf{RQ3 (modality):} considering multi-modal encoders, \textit{which inputs most improve flood and permanent-water discrimination (e.g., optical, RTC, GRD)?}

    \item \textbf{RQ4 (generalization):} \textit{does training on \dataset{} transfer to unseen,
    out-of-period events better than existing benchmarks?}
\end{itemize}

In an attempt to answer these questions, we provide three contributions: (1)~\datasetns{}, a global flood benchmark dataset with a dedicated permanent water layer and multi-sensor co-registration at scale; (2)~a reproducible training and evaluation protocol, from single-image binary segmentation to multi-temporal multi-class segmentation; and (3)~an extensive backbone benchmark for flood detection and water body segmentation, comparing foundation models against conventional backbones and \dataset{}against existing datasets.

%% file: sections/2_relworks.tex
\section{Related Work}
\label{sec:related}
\vspace{-0.1cm}
\subsubsection{Flood Mapping Datasets.}
\label{sec:related:datasets}

\input{tables/dataset_comparison}

Considering flood-specific datasets, there is little shared consensus on input modalities or temporal modelling. Optical datasets such as \emph{WorldFloods}~\cite{mateogarcia2021worldfloods,portalesjulia2023worldfloods} lack SAR entirely, while SAR-centric datasets such as \emph{MMFlood}~\cite{montello2022mmflood}, \emph{Kuro~Siwo}~\cite{bountos2023kurosiwo}, and \emph{S1GFloods}~\cite{saleh2024damnet} provide no optical counterpart, and \emph{Sen1Floods11}~\cite{bonafilia2020sen1floods11} pairs the two only at a single flood-time acquisition, precluding bi-temporal change analysis. Datasets that do combine SAR and optical (\emph{OmbriaNet}~\cite{drakonakis2022ombrianet}, \emph{CAU-Flood}~\cite{he2023cross}, and \emph{STURM-Flood}~\cite{notarangelo2025sturm}) are, however, missing pre-event SAR or remain largely unpaired (CAU-Flood pairs pre-event S2 with post-event S1 only; STURM-Flood supplies 21\,602 S1 tiles but only 2\,675 corresponding S2 images). Furthermore, none of them distinguishes flooding from permanent water.

\emph{Annotation strategies} vary widely. Existing labels range from automatically derived or weakly supervised masks (i.e., prone to systematic noise, especially from optical composites under flood-time cloud cover) to fully manual delineations, as in \emph{Kuro~Siwo}, re-annotated by SAR specialists, or \emph{Sen1Floods11}, which hand-labels only 446 of its 4\,831 tiles (${<}10\%$). Manual annotation, however, is not by itself a guarantee of quality: as \cref{fig:dataset_comparison_visual} shows, hand-drawn delineations do not necessarily yield cleaner or more accurate boundaries than well-curated automated products.

\emph{SAR processing heterogeneity} is a further issue: datasets distribute SAR at different processing levels and value ranges (e.g., raw GRD \cite{montello2022mmflood}, terrain-corrected RTC \cite{bountos2023kurosiwo}) with no common normalization, hindering cross-dataset evaluation and operational deployment.
\emph{Split design} is also often overlooked. Several datasets adopt random tile-level splits that risk spatial leakage between train and test~\cite{drakonakis2022ombrianet,notarangelo2025sturm,rahnemoonfar2021floodnet}, possibly compromising reported performance.

Finally, \emph{permanent water derivation} is far from standardized. MMFlood adopts OpenStreetMap hydrography \cite{montello2022mmflood}, although incomplete, while the JRC Global Surface Water product~\cite{pekel2016high} used by WorldFloods and Sen1Floods11 is Landsat-derived at \SI{30}{\metre} and lacks most narrow rivers and small water bodies. In both cases, these systematic gaps propagate into the flood label. \Cref{tab:dataset_comparison} summarizes the properties of existing flood datasets.

\dataset is designed to address these limitations jointly. It pairs fully co-registered pre- and post-event Sentinel-1 with a cloudless pre-event Sentinel-2 composite as optical reference; distributes SAR at two standardized processing levels (GRD and RTC), removing the cross-dataset normalization gap; draws its ground truth from manually curated CEMS delineations that prioritize SAR-derived sources, sidestepping cloud-induced optical noise; derives a dedicated permanent water layer from AlphaEarth Foundations embeddings\cite{alphaearth} to separate permanent from flooded water; and adopts event-level partitioning with a temporally disjoint held-out test set to rule out spatial leakage.

\vspace{-0.4cm}
\subsubsection{Flood Delineation Methods.}
\label{sec:related:methods}
Classical approaches to SAR-based flood delineation apply \emph{intensity thresholding} or fuzzy-logic rules to exploit the low backscatter signature of open water~\cite{martinis2009,twele2016}.
These methods remain competitive baselines on small, homogeneous scenes but degrade in urban areas, dense vegetation, and turbulent water conditions.
Deep learning segmentation has largely replaced classical methods on benchmark tasks.
Fully convolutional networks applied to SAR data~\cite{kang2018flood} and encoder-decoder architectures based on U-Net~\cite{ronneberger2015unet} and DeepLabV3+~\cite{chen2018deeplabv3} dominate current flood-segmentation benchmarks.
Change detection architectures, including dual-branch transformers~\cite{bandara2022changeformer} and Siamese networks applied to SAR pairs~\cite{zhao2023siamdwe}, have demonstrated strong performance in flood detection tasks.
The recent success of geospatial foundation models has extended transferable representations to many downstream tasks including land-cover mapping, crop monitoring, and disaster assessment.
Models such as OlmoEarth~\cite{herzog2025olmoearth}, DOFA~\cite{xiong2024dofa}, and TerraMind~\cite{jakubik2025terramind} have demonstrated strong generalization across different sensors and areas.
Their growing adoption and increasing number of different foundation models and approaches have exposed the need to rigorously benchmark these models on complex tasks, with large-scale, robust datasets \cite{doerksen2026earthshift}. 

%% file: tables/dataset_comparison.tex
\begin{table*}[t]
  \caption{Comparison of satellite-based flood segmentation datasets.\\
  { JRC-GSW: JRC Global Surface Water; OSM: OpenStreetMap; HL: hand-labeled; AEF: AlphaEarth-derived.} { $^\star$Image size reported as the average size of raw tiles.}}
  \label{tab:dataset_comparison}
  \centering
  \resizebox{\linewidth}{!}{
      \begin{tabular}{@{}lccccccccc@{}}
        \toprule
        Dataset
          & Modality
          & Resolution
          & Image size
          & Events
          & Tiles
          & Area (km$^2$)
          & Temporal
          & Coverage
          & Perm.\ water \\
        \midrule
        FloodNet~\cite{rahnemoonfar2021floodnet}
          & RGB (drone)
          & VHR
          & $4\,000\times3\,000$
          & 1
          & 2\,343
          & N/A
          & post
          & 2017
          & -- \\
        Sen1Floods11~\cite{bonafilia2020sen1floods11}
          & S1, S2
          & \SI{10}{\metre}
          & $512\times512$
          & 11
          & 4\,831
          & 120\,406
          & post
          & 2016--2019
          & JRC-GSW, HL \\
        ETCI 2021~\cite{etci2021}
          & S1
          & \SI{10}{\metre}
          & $256\times256$
          & 5
          & 33\,405
          & 36\,623
          & post
          & 2017--2019
          & N/A \\
        WorldFloods v2~\cite{portalesjulia2023worldfloods}
          & S2
          & \SI{10}{\metre}
          & $3\,443\times3\,204^\star$
          & 144
          & 509
          & 586\,618
          & post
          & 2016--2023
          & JRC-GSW \\
        MMFlood~\cite{montello2022mmflood}
          & S1, DEM
          & \SI{20}{\metre}
          & $909\times993^\star$
          & 95
          & 1\,748
          & 137\,659
          & post
          & 2014--2021
          & OSM hydro. \\
        OmbriaNet~\cite{drakonakis2022ombrianet}
          & S1, S2
          & \SI{10}{\metre}
          & $256\times256$
          & 23
          & 844
          & 553
          & pre+post
          & 2017--2021
          & CEMS \\
        Kuro Siwo~\cite{bountos2023kurosiwo}
          & S1, DEM
          & \SI{10}{\metre}
          & $224\times224$
          & 43
          & 67\,490
          & 338\,000
          & pre+post
          & 2015--2022
          & HL \\
        CAU-Flood~\cite{he2023cross}
          & S1, S2
          & \SI{10}{\metre}
          & $256\times256$
          & 18
          & 18\,302
          & 95\,142
          & pre+post
          & 2016--2022
          & HL \\
        S1GFloods~\cite{saleh2024damnet}
          & S1
          & \SI{10}{\metre}
          & $256\times256$
          & 46
          & 5\,360
          & 35\,127
          & pre+post
          & 2015--2022
          & HL \\
        \multirow{2}{*}{STURM-Flood~\cite{notarangelo2025sturm}}
          & S1
          & \multirow{2}{*}{\SI{10}{\metre}}
          & \multirow{2}{*}{$128\times128$}
          & 47 
          & 21\,602
          & \multirow{2}{*}{35\,399}
          & \multirow{2}{*}{post}
          & \multirow{2}{*}{2016--2024}
          & \multirow{2}{*}{OSM} \\
          & S2 & & & 29 & 2\,675 & & & & \\
        \midrule
        \textbf{\datasetns} \hspace{2px}
          & S1, S2, DEM
          & \SI{10}{\metre}
          & $1\,024\times1\,024$
          & 219
          & 14\,282
          & 1\,141\,749
          & pre+post
          & 2016--2026
          & AEF \\
        \bottomrule
      \end{tabular}
  }
\end{table*}

%% file: sections/3_dataset.tex
\section{The \dataset Dataset}
\label{sec:dataset}
\vspace{-0.1cm}
\dataset is a large-scale, multi-modal benchmark for flood segmentation, pairing co-registered Sentinel-1 SAR and Sentinel-2 optical imagery with manually filtered flood masks derived from CEMS Rapid Mapping activations. It spans 219 flood events across 65 countries and a decade of acquisitions, capturing a diversity of climates, land cover, and sensor conditions absent from existing flood datasets. We detail the data sources, the construction pipeline, and the resulting statistics in the next sections.

\subsection{Data sources}
\label{sec:dataset:sources}

Each CEMS activation corresponds to a flood event and contains one or more Areas of Interest (AoIs), regions impacted by the event. Large-scale events can have dozens of AoIs of heterogeneous sizes and shapes.
For every pair of event and AoI, CEMS publishes a series of vector delineation products over the hours and days following the disaster, progressively refining the mapped flood extent as new satellite acquisitions become available. Each product is associated with a pre-event \emph{reference image}, used to assess the situation before the disaster, and a post-event image, which depicts the situation after the flood. 
We select a single product per pair: we prioritize products derived from Sentinel-1 and Sentinel-2 imagery, so that annotation and training data share the same sensor and resolution; when no Sentinel-derived product exists, we select the first available product with the highest quality, following CEMS directives \cite{cems} (in decreasing order: \textit{Grading}, \textit{Delineation Monitoring}, \textit{Delineation}, \textit{First Estimate}).
All selected labels were manually inspected and corrected where necessary.

Each product carries three reference dates: the \emph{event date} (when the flood occurred), the \emph{pre-event image date}, and the \emph{post-event image date}, the latter two being the acquisition dates of the satellite scenes used to produce the final analysis. It also records a \emph{sensor} field, namely the satellite from which each acquisition was derived.
For each event-AoI pair, the recorded date and sensor determine which acquisition we retrieve: if the sensor is Sentinel-1 and the exact post-event scene used by CEMS analysts is available, we retrieve that same scene; otherwise, we take the closest available acquisition after the event date.
We mirror this procedure for the pre-event acquisition, taking the image corresponding to the reported pre-event date, or nearest acquisition preceding the event date instead.

Four satellite sources are co-registered at \SI{10}{\metre} ground sampling distance: Sentinel-1 Ground Range Detected (GRD) and Radiometrically Terrain-Corrected (RTC) products (VV and VH polarizations, pre- and post-event), a pre-event Sentinel-2 composite (Level-2A surface reflectance, 12 spectral bands resampled to \SI{10}{\metre}), and the Copernicus GLO-30 DEM~\cite{glo30} as a static elevation layer.

\subsection{Construction pipeline}
\label{sec:dataset:pipeline}

Dataset construction comprises five automated and reproducible stages operating directly on public Copernicus products:
\vspace{-0.2cm}
\paragraph{Spatial partitioning.} Using metadata from each CEMS activation, we subdivide each variable AoI into regular \SI{10240}{\metre} square bounding boxes aligned to UTM grids, giving a consistent spatial footprint across events.
\vspace{-0.2cm}
\paragraph{Data retrieval.} For each AoI we retrieve Sentinel-1 GRD and RTC at both the pre- and post-event dates through the Sentinel Hub APIs~\cite{sentinelhub}. Sentinel-2 is retrieved for the pre-event period only: floods are typically accompanied by persistent cloud cover which, compounded by the optical revisit interval, makes a clear acquisition near the post-event delineation date extremely unlikely \cite{portalesjulia2023worldfloods} (see \cref{fig:dataset_comparison_visual}). Given the optical source, we minimize the cloud coverage by applying a median composite over each tile, selecting a window of three weeks from the event date, and a maximum of 3 S2-L2A acquisitions. Considering terrain, we download the Copernicus GLO-30 DEM and resample it to the \SI{10}{\metre} grid as an additional static layer.
\vspace{-0.2cm}
\paragraph{Validity mask generation.} We derive two validity masks. Despite the median composite, certain geographical areas may still display cloud coverage. For this reason, an auxiliary \emph{cloud mask} is produced by running OmniCloudMask~\cite{wright2025omnicloudmask} on the pre-event Sentinel-2 scene, yielding per-pixel \textit{clear}, \textit{thin}, or \textit{thick} labels. We further generate a pixel \emph{validity mask} that marks usable pixels as the intersection of the AoI boundary, the image footprint, and the tile bounding box.
\vspace{-0.2cm}
\paragraph{Label composition.} CEMS products map flood extent but not permanent water; since single-image water segmentation must distinguish the two, separating these classes is a core design decision of \datasetns.
As existing global layers are unsuitable at our resolution (\cref{sec:related:datasets}), we provide a dedicated \SI{10}{\metre} permanent water layer by training a lightweight model on the \emph{Earth Surface Water} (ESW) dataset~\cite{luo2021esw} from annual \emph{AlphaEarth Foundations} (AEF) embeddings~\cite{alphaearth} (full details and examples in \cref{sec:permwater}). We rasterize into flood labels only those CEMS polygons explicitly categorized as flood, excluding trace-level annotations. The final label merges these layers into \textit{background}, \textit{permanent water}, \textit{flooded water}, and \textit{invalid}, the last assigned to pixels under thick cloud when the cloud mask is selected, or outside the validity mask.
\vspace{-0.2cm}
\paragraph{Quality filtering.} We discard bounding boxes that had missing or partial modalities, excessive cloud cover, or imagery inconsistent with the reference label; this is common in flash floods, where even a small acquisition-to-delineation gap misaligns annotations.

\subsection{Dataset statistics}
\label{sec:dataset:stats}

\begin{figure}[t]
  \centering
  \includegraphics[width=\linewidth]{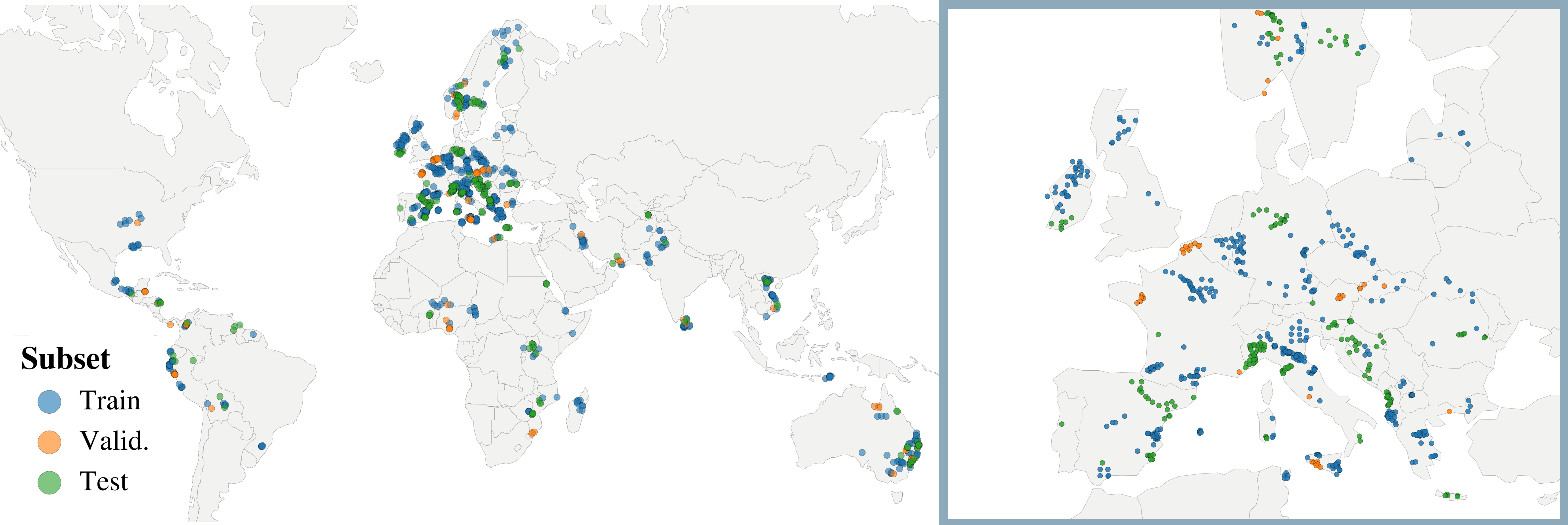}
  \caption{Global distribution of \dataset AoIs, colored by split assignment (train/validation/test). The inset enlarges Europe, where touching AoIs share a split to prevent boundary leakage.}
  \label{fig:split_map}
\end{figure}

\dataset covers 219 flood events spanning January 2016 to March 2026, of which the most recent form a temporally disjoint held-out set reserved for cross-dataset experiments.
The dataset spans 65 countries across six continents, with a minimum of 13 up to a maximum of 30 events per year.
The events decompose into 1\,055 valid event-AoI pairs, derived from applying the quality filtering stage described above to a pool of 1\,333 candidate areas.
Each pair may yield one or more \SI{10240}{\metre} bounding boxes, producing a total of 14\,282 valid tiles at $1024\times1024$ pixels.
Every tile provides a complete (S1\textsubscript{pre}, S1\textsubscript{post}, S2\textsubscript{pre}, DEM) tuple of curated, ML-ready imagery.
Given the source catalog, Europe dominates the geographic distribution with 140 events, as shown in \cref{fig:split_map}.
However, several large-scale events have been mapped across the globe, and we deliberately ensured that regions outside Europe remain well represented across splits.

\begin{figure}[t]
  \centering
  \includegraphics[width=\linewidth]{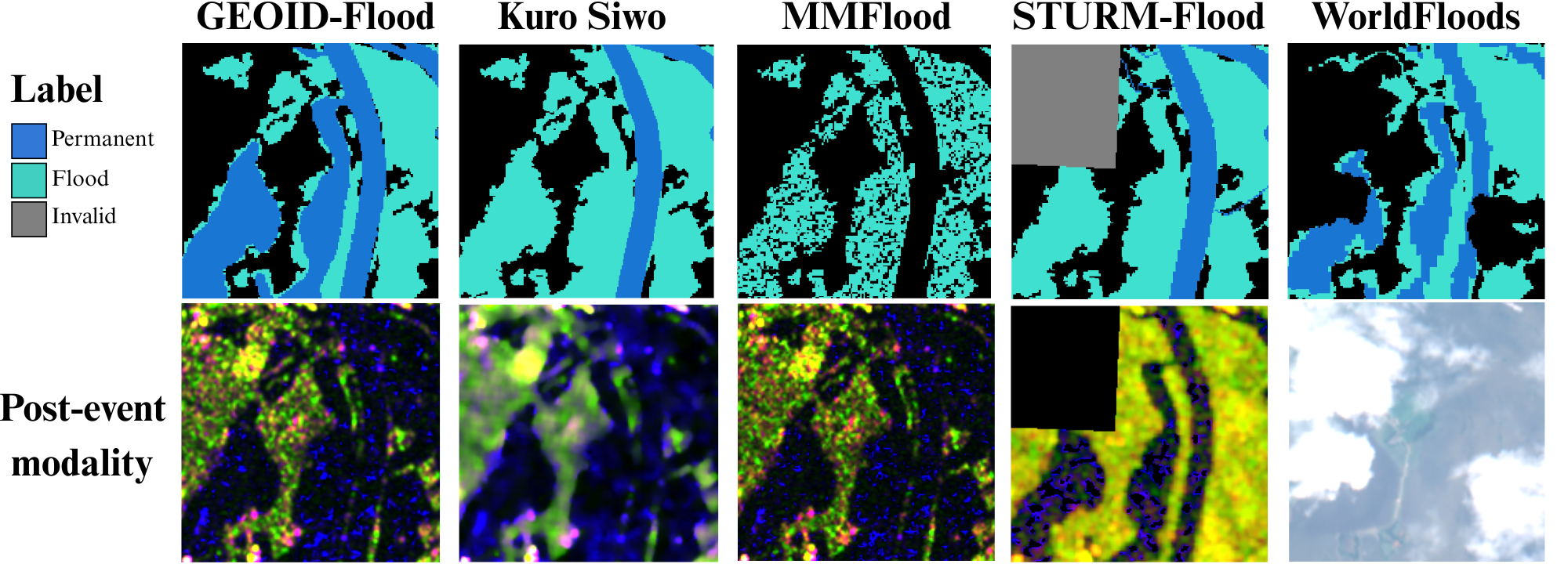}
  \caption{Visual comparison of \dataset against other popular datasets on a shared area. First row: available labels, second row: corresponding post-event modality of each dataset.
  }
  \label{fig:dataset_comparison_visual}
\end{figure}

Specifically, AoIs are stratified by continent and sampled with target proportions of 70/10/20\,\% for training, validation and test respectively, so that every region is proportionally represented in each subset.
To prevent boundary leakage, adjacent or overlapping AoIs are constrained to the same split, keeping spatially contiguous areas together (see \cref{fig:split_map}).
This event-level partitioning yields 8\,938 tiles for training, 1\,241 for validation, and 2\,674 for testing. A further 1\,429 tiles, drawn from events post-dating January 2026, form the temporally disjoint held-out set, used exclusively for the cross-dataset comparison of \cref{sec:results:comparison}.
\Cref{fig:dataset_overview} shows representative tiles from three \dataset events with all modalities and label layers; \cref{fig:dataset_comparison_visual} gives a side-by-side comparison with literature datasets on a shared event.

\vspace{-0.2cm}
\subsubsection{Held-out set}
\label{sec:dataset:crossdataset}
To support cross-dataset generalization experiments, we construct a dedicated held-out test set from CEMS activations published after January 2026 (EMSR857--EMSR871, 83 event-AoI pairs, spanning February--March 2026).
The chosen window (February--March 2026) provides a sufficient number of activations to be used as a test set while remaining temporally disjoint, by construction, from our own splits and all the other datasets mentioned in \cref{sec:related:datasets}.
The held-out set follows the same construction pipeline and modality structure as the main dataset, and is used exclusively for the cross-dataset experiments in \cref{sec:results:comparison}.

%% file: sections/4_method.tex
\section{Methodology}
\label{sec:method}
\vspace{-0.1cm}
We benchmark geospatial foundation models and conventional encoders on \dataset under a
shared training protocol (\cref{sec:suppl:optim}), aimed at answering the research questions of \cref{sec:intro}.
After defining the segmentation task (\cref{sec:method:task}), we organize the experiments into three training scenarios of increasing complexity (\cref{sec:method:paradigms}): a
single-image backbone benchmark (\textbf{RQ1}); paired training that introduces explicit flood supervision (\textbf{RQ2}); and fusion that adds optical context (\textbf{RQ3}). Two further studies, reported in \cref{sec:results}, complete the picture: a modality ablation that
isolates input contributions (\textbf{RQ3}, \cref{sec:results:modality}) and a cross-dataset
protocol that measures generalization to unseen events (\textbf{RQ4}, \cref{sec:results:comparison}).

\subsection{Task formulation}
\label{sec:method:task}

We formulate flood mapping as a per-pixel semantic segmentation problem. Each tile carries a
three-class label: background, permanent water, and flooded water. We evaluate models in both
\emph{single-image} and \emph{multi-image} settings, each with its own target formulation.
A single SAR acquisition does not, in general, provide enough evidence to separate flooded
water from permanent water, as both yield similarly dark returns in VV/VH backscatter.
For single-image training we therefore reduce the problem to a binary \emph{water-body
segmentation}, remapping labels according to the acquisition time step: on pre-event tiles,
flooded pixels are relabelled as background, since inundation is only meaningful after the
event; on post-event tiles, flooded and permanent water are merged into a single water class,
so the target reflects total surface-water extent. When pre- and post-event images are
processed jointly, temporal context makes the two water classes separable, and the full
three-class target is retained.
In all settings, pixels outside the CEMS analysis area and other invalid pixels are excluded from the loss.

\subsection{Training scenarios}
\label{sec:method:paradigms}

The single-image and multi-image settings of \cref{sec:method:task} instantiate as three
scenarios of increasing temporal and modal complexity. Scenario~(i) is run across the full
encoder zoo (\cref{sec:suppl:zoo}) as our backbone comparison; since scenarios~(ii)--(iii)
probe temporal and multi-modal design choices rather than the backbone itself, we fix the
encoder there to a single backbone selected from~(i).
\vspace{-0.2cm}
\paragraph{(i) Single-image.} Pre- and post-event crops are treated as independent samples:
the model performs one forward pass per tile, trained with cross-entropy (CE) on the remapped
binary water-body labels from \cref{sec:method:task}. This is our primary instrument for
ranking frozen and finetuned foundation models against conventional ImageNet-pretrained
encoders (\textbf{RQ1}); its binary predictions also form the post-hoc three-class baseline
against which explicit flood modelling is measured (\textbf{RQ2}).
\vspace{-0.2cm}
\paragraph{(ii) Paired, two-pass.} We draw co-registered pre-/post-event pairs and apply the same encoder-decoder to each time step in two separate forward passes, summing the two CE
terms under the binary remapping of~(i). To this we add a \emph{flood-change} loss (\textbf{RQ2}): a binary CE term on pixels that are water post-event but not pre-event, sharpening sensitivity to the flooded-water class. We further vary the \emph{pre-event modality} (\textbf{RQ3}): the default \emph{S1\,$\rightarrow$\,S1} pairing is compared against \emph{S2\,$\rightarrow$\,S1} (optical pre-event, radar post-event) and \emph{S1\,+\,S2\,$\rightarrow$\,S1} (both pre-event modalities), testing whether pre-event optical context helps.

\vspace{-0.2cm}
\paragraph{(iii) Paired, single-pass (fusion).} Both images enter in a single forward pass and
the target retains its full three-class structure (\textbf{RQ2}). We compare two fusion
strategies: \emph{early fusion} stacks the two acquisitions along the channel dimension through
one encoder-decoder, while \emph{mid fusion} encodes each in a separate branch and merges the
feature maps by element-wise subtraction (post minus pre) before a shared decoder. Each
strategy is run on three pre-/post-event pairings: Sentinel-1 GRD alone (\emph{S1}), the
pre-event Sentinel-2 image with post-event SAR (\emph{S2\,$\rightarrow$\,S1}), and an
optically augmented stack (\emph{S1\,+\,S2}) that concatenates the two modalities.
Contrasting~(iii) with~(i)--(ii) (\cref{sec:results:eval}) isolates the benefit of explicit over
post-hoc flood modelling (\textbf{RQ2}).

%% file: sections/5_results.tex
\section{Experiments}
\label{sec:results}

\subsection{Evaluation protocol}
\label{sec:results:eval}

All models and scenarios are evaluated on the test split (\cref{sec:dataset:stats}) under two
tasks: a generic binary \textit{water-body segmentation} (water vs.\ background, on remapped
labels) and a specific multiclass \textit{flood detection} (background, permanent water,
flooded water), using a standard U-Net decoder in every configuration.
For single-image and paired models (scenarios~(i) and~(ii)), the three-class map is derived at
inference without a dedicated head: we run the binary model on the pre- and post-event tiles,
combine the two water masks, and assign flood as their pre-/post-event difference. Scenario~(iii)
predicts the three classes directly. We report both tasks in F1 score and Intersection over
Union (IoU), where the subscript \textit{avg} indicates macro-averaged results, and take
IoU$_{\mathrm{flood}}$ as the primary reference metric for flood detection. Throughout, we use
\textit{binary} for water-body delineation results and \textit{multiclass} for flooded-area
delineation.

\subsection{Model benchmark}
\label{sec:results:backbone}

\begin{figure*}[t]
  \centering
  \includegraphics[width=\linewidth]{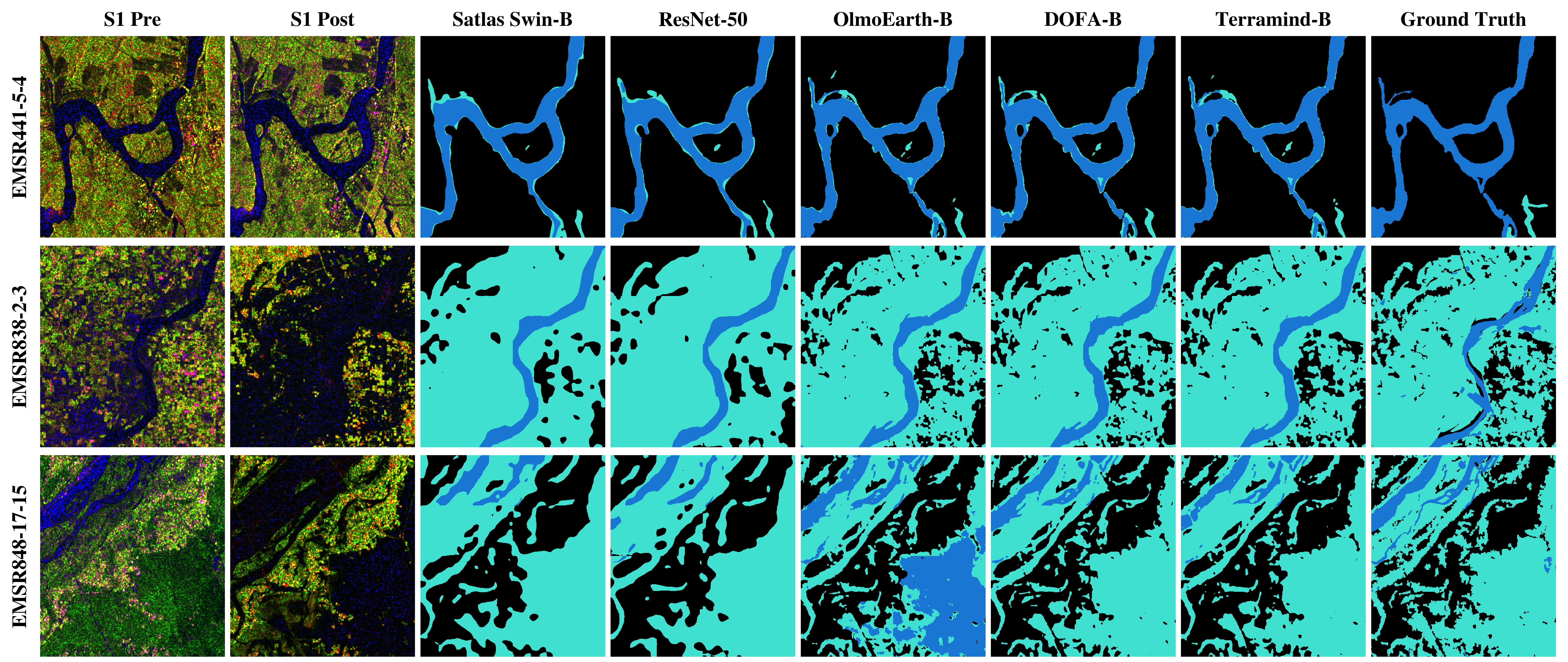}
  \caption{Qualitative comparison on three \dataset test events (rows) 
  (\cref{sec:method:paradigms}), using five different encoders, finetuned under scenario~(i)}
  \label{fig:qualitative_backbone}
\end{figure*}

\Cref{tab:backbone_benchmark} addresses RQ1 across the full encoder zoo under scenario~(i), while \cref{fig:qualitative_backbone} shows some inferences on the test set. Both show how narrow the gap is: all models but the frozen Satlas Swin-B fall within a 0.04
range of binary IoU (0.844--0.884). The finetuned version of \textit{TerraMind-L} reaches the best results
(IoU$_{\mathrm{bin}}$ 0.884, F1$_{\mathrm{bin}}$ 0.936). However, \textit{Swin-T}  reaches 0.873 IoU while being nearly an order of magnitude smaller (32\,M vs.\ 323\,M), matching or outperforming every finetuned foundation model except TerraMind in \textit{base} and \textit{large} variants.
The gap between geospatial foundation models and ImageNet-pretrained encoders is therefore small under this shared protocol: with a strong shared decoder and adequate training, the backbone is not the bottleneck, and most encoders converge to similar scores, leaving
remote sensing-specific pre-training a consistent but modest edge on SAR water segmentation.
Finetuning gives small gains to already robust backbones, but provides sizable gains for weaker or smaller ones (e.g., Satlas Swin-B, 0.751\,$\rightarrow$\,0.861). While binary water segmentation is well handled across the benchmark, the \textit{flood} class remains harder to tackle, with the highest IoU$_{\mathrm{flood}}$ at 0.484. This motivates the need for flood-specific approaches, presented in the following sections.
Guided by the benchmark, we adopt \textit{TerraMind-B} for all remaining experiments, given its balance between accuracy and practicality for operational use: negligible loss in performance w.r.t. the top performing model, at roughly a third of the parameters (101\,M vs.\ 323\,M).

\input{tables/backbone_benchmark}

\subsection{Paired and change-focused extensions}
\label{sec:results:paired}

\begin{figure*}[t]
  \centering
  \includegraphics[width=\linewidth]{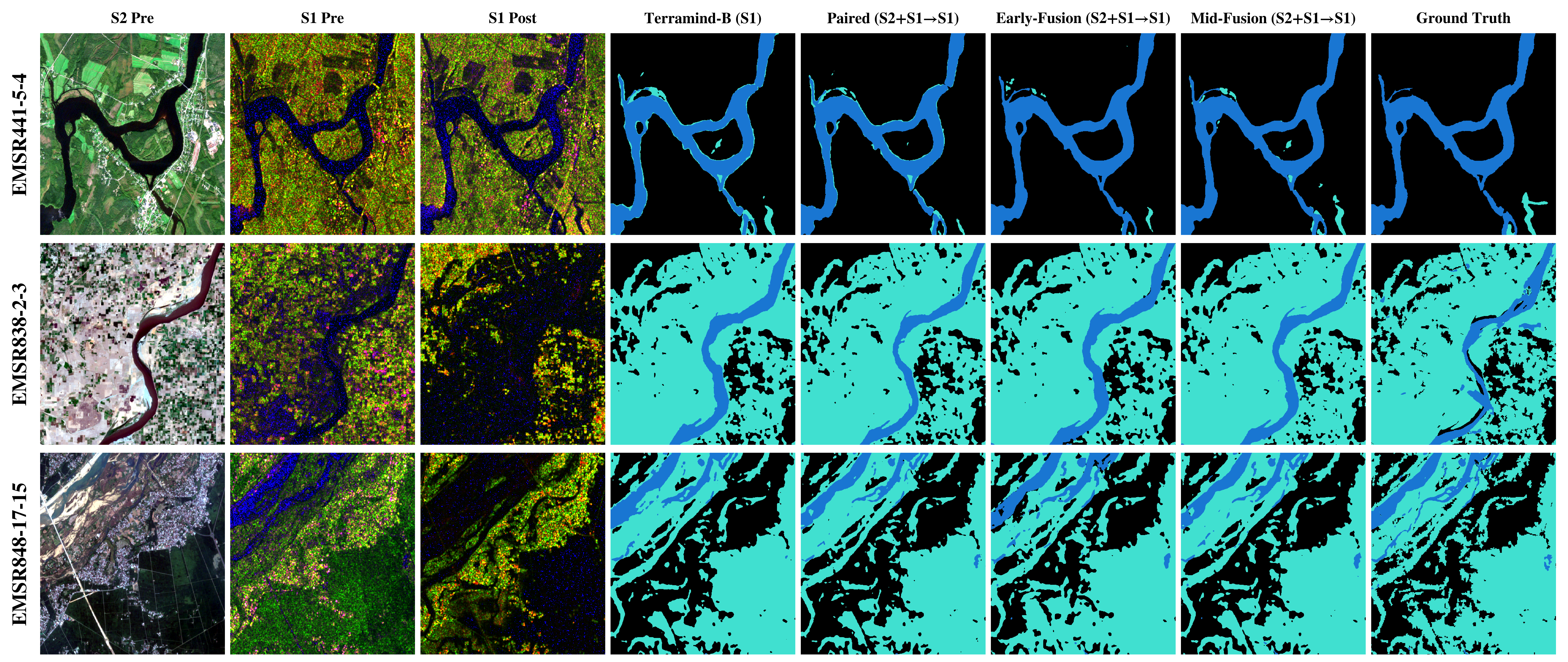}
  \caption{Qualitative comparison on three \dataset test events (rows)
  (\cref{sec:method:paradigms}) across the scenarios of \cref{tab:paired_multimodal}: the single-image baseline (i, \textit{TerraMind-B} on S1), the paired double-pass model (ii), and early- and mid-fusion (iii).}
  \label{fig:qualitative_scenarios}
\end{figure*}

Focusing now on RQ2, we investigate whether temporal pairing improves on the single-image
baseline, and whether explicitly modelling the flood change can further improve performance.
\Cref{tab:paired_multimodal} reports a progression of increasingly explicit flood modelling approaches, from single-image differencing (scenario~(i)), to paired options in double and single pass (scenarios~(ii)--(iii)), while \cref{fig:qualitative_scenarios} shows representative inferences across these scenarios on the test set.

For scenario~(ii), the paired flood-change loss matches the post-hoc baseline on binary water when frozen but trails it on flooded water; finetuning lifts IoU$_{\mathrm{flood}}$ only marginally over the baseline (0.486 vs.\ 0.479).
Considering scenario~(iii), ad-hoc feature fusion is a viable approach, but only when finetuned. Both fusion variants sit near or below the baseline when frozen; early fusion suffers most, as its single shared encoder must represent both acquisitions identically, whereas mid fusion may benefit from separate branches. Finetuning lifts both above the baseline, suggesting that the gain derives from learning the change end-to-end rather than from pairing.

Anticipating RQ3, adding optical context proves decisive. Finetuned early fusion reaches the best flooded-water IoU (0.521) and binary water (F1$_{\mathrm{bin}}$ 0.942), and how the optical is included has little effect on results: using the pre-event Sentinel-2 composite in place of the pre-event SAR (\emph{S2\,$\rightarrow$\,S1}) matches stacking it onto the bi-temporal SAR (\emph{S1\,+\,S2}).

\input{tables/paired_multimodal}

\subsection{Modality ablation}
\label{sec:results:modality}

Given the available multi-modal encoders and the fusion results, we investigate which inputs
are most effective in this context (RQ3). Isolating a frozen TerraMind-B, we vary only the
input stack across the four available modalities (S1-GRD, S1-RTC, S2-L2A, DEM) in different
combinations (\cref{tab:modality_ablation}). Since Sentinel-2 is unavailable post-event, we
restrict the ablation to pre-event tiles, where flooded water is absent and the task reduces to water-body delineation.
The Sentinel-1 product is not as influential as one may think; however, \textit{GRD} consistently edges out \textit{RTC}
(IoU$_{\mathrm{bin}}$ 0.931 vs.\ 0.922), suggesting that terrain backscatter correction might not matter in this case, and resampling might even introduce slight artifacts.
Likewise, the DEM stays within noise of the baseline (0.931\,$\rightarrow$\,0.934, within $\pm$0.003 std), indicating that raw elevation data might be redundant when paired with the SAR signal. Pre-event Sentinel-2 instead helps most by a wide margin (+0.015, to 0.946), enriching features with optical information.
Nevertheless, we do not rule out that purpose-built or more recent encoders could better exploit all these modalities; we therefore retain them in the released dataset to support future work.

\input{tables/modality_ablation}

\subsection{Cross-dataset generalization}
\label{sec:results:comparison}

Finally, we assess whether training on \dataset generalizes to unseen events better than
existing benchmarks (RQ4). We train the same U-Net (TerraMind-B), frozen and finetuned, on each
of Kuro~Siwo~\cite{bountos2023kurosiwo}, MMFlood~\cite{montello2022mmflood},
WorldFloods\,v2~\cite{portalesjulia2023worldfloods}, and
Sen1Floods11~\cite{bonafilia2020sen1floods11}, and evaluate every model on the held-out set
(\cref{sec:dataset:crossdataset}). The external datasets follow widely different SAR
preprocessing conventions (\cref{fig:dataset_comparison_visual}); to remove this as a
confounder, we re-download and reprocess every scene through a common Sentinel-1 RTC pipeline
(terrain-flattened $\gamma^0$, GLO-30 DEM, \SI{10}{\metre}; \cref{sec:dataset:sources}) and
apply the same channel normalization throughout. Sentinel-1 sources are re-acquired at their
original dates, while WorldFloods\,v2, annotated on Sentinel-2, is paired with its temporally
closest Sentinel-1 scene, dropping pairs more than two days apart. Because preprocessing and
normalization are shared and no model has seen any held-out event, models differ only in the
training set they learned from.

\input{tables/cross_dataset_comparison}

\dataset is the strongest training source in both regimes. On binary water it transfers best, ahead of every external source including the
CEMS-derived Kuro~Siwo, and the three-class metrics follow the same ordering. Among external sources, Kuro~Siwo transfers best on
binary water and WorldFloods\,v2 on the three-class task (F1$_{\mathrm{avg}}$ 0.816), the latter
notable given its optical-derived labels.
Finetuning the source does not overturn this ordering and is dataset-dependent: it clearly helps
Sen1Floods11 but destabilizes MMFlood, whose strong flood/non-flood imbalance makes full finetuning harder (IoU$_{\mathrm{flood}}$ 0.512$\rightarrow$0.380).

Since the held-out permanent water labels share \dataset's derivation, the binary water lead is partly expected; the flood metrics, independently derived from CEMS, confirm that scale and diversity yield genuine transfer gains.

%% file: tables/backbone_benchmark.tex
\begin{table*}[t]
  \centering
  \caption{Single-modality (Sentinel-1 GRD) benchmark on the \dataset test split.
  All models are trained under scenario (i) (\cref{sec:method:paradigms}).
  Metrics follow \cref{sec:results:eval}.}
  \label{tab:backbone_benchmark}
  {\footnotesize
  \setlength{\tabcolsep}{6pt}
  \renewcommand{\arraystretch}{1.05}
  \setlength{\aboverulesep}{0pt}
  \setlength{\belowrulesep}{0pt}
  \resizebox{\textwidth}{!}{%
  \begin{tabular}{@{}ll|cc|ccc|cc@{}}
    \toprule
    & & \multicolumn{2}{c|}{\textbf{Binary}} & \multicolumn{5}{c}{\textbf{Multiclass}} \\
    \textbf{Model} & \textbf{Params (M)} & \textbf{F1} & \textbf{IoU} & \textbf{IoU}$_\mathrm{bg}$ & \textbf{IoU}$_\mathrm{perm}$ & \textbf{IoU}$_\mathrm{flood}$ & \textbf{IoU}$_\mathrm{avg}$ & \textbf{F1}$_\mathrm{avg}$ \\
    \midrule
    \midrule
    \multicolumn{9}{l}{\textit{Frozen Encoder Foundation Models}} \\
    \midrule
    TerraMind-T      & 12.33 (6.90)   & 0.926 & 0.868 & 0.974 & 0.823 & 0.460 & 0.752 & 0.840 \\
    TerraMind-S      & 30.92 (9.44)   & 0.924 & 0.866 & 0.973 & 0.827 & 0.448 & 0.749 & 0.837 \\
    TerraMind-B      & 100.87 (15.53) & 0.931 & 0.877 & 0.975 & 0.849 & 0.478 & 0.767 & 0.851 \\
    TerraMind-L      & 322.83 (20.32) & \underline{0.934} & \underline{0.881} & \underline{0.976} & 0.856 & \textbf{0.484} & \underline{0.772} & \underline{0.854} \\
    DOFA-B           & 126.92 (15.53) & 0.924 & 0.864 & 0.973 & 0.797 & 0.450 & 0.740 & 0.831 \\
    DOFA-L           & 357.53 (20.32) & 0.921 & 0.861 & 0.972 & 0.792 & 0.444 & 0.736 & 0.828 \\
    OlmoEarth-B      & 104.48 (15.53) & 0.925 & 0.867 & 0.974 & 0.808 & 0.445 & 0.742 & 0.832 \\
    SSL4EO (RN-50)   & 37.98 (12.43)  & 0.910 & 0.844 & 0.969 & 0.771 & 0.413 & 0.718 & 0.813 \\
    Satlas Swin-B    & 92.72 (12.27)  & 0.842 & 0.751 & 0.952 & 0.592 & 0.213 & 0.586 & 0.690 \\
    \midrule
    \midrule
    \multicolumn{9}{l}{\textit{Finetuned Encoder Foundation Models}} \\
    \midrule
    TerraMind-T      & 12.33  & 0.926 & 0.869 & 0.974 & 0.823 & 0.461 & 0.753 & 0.840 \\
    TerraMind-S      & 30.92  & 0.926 & 0.868 & 0.973 & 0.824 & 0.469 & 0.756 & 0.843 \\
    TerraMind-B      & 100.87 & 0.932 & 0.878 & 0.975 & \underline{0.856} & 0.479 & 0.771 & 0.853 \\
    TerraMind-L      & 322.83 & \textbf{0.936} & \textbf{0.884} & \textbf{0.977} & \textbf{0.871} & 0.478 & \textbf{0.775} & \textbf{0.855} \\
    DOFA-B           & 126.92 & 0.929 & 0.873 & 0.975 & 0.833 & 0.477 & 0.761 & 0.847 \\
    DOFA-L           & 357.53 & 0.928 & 0.872 & 0.974 & 0.822 & 0.471 & 0.756 & 0.843 \\
    OlmoEarth-B      & 104.48 & 0.926 & 0.868 & 0.973 & 0.830 & \underline{0.484} & 0.757 & 0.845 \\
    SSL4EO (RN-50)   & 37.98  & 0.920 & 0.858 & 0.972 & 0.813 & 0.428 & 0.737 & 0.827 \\
    Satlas Swin-B    & 92.72  & 0.921 & 0.861 & 0.972 & 0.813 & 0.449 & 0.745 & 0.834 \\
    \midrule
    \midrule
    \multicolumn{9}{l}{\textit{ImageNet-Pretrained Supervised Backbones}} \\
    \midrule
    ResNet-50        & 29.34  & 0.919 & 0.857 & 0.971 & 0.808 & 0.431 & 0.737 & 0.827 \\
    ResNet-101       & 48.33  & 0.916 & 0.853 & 0.971 & 0.800 & 0.421 & 0.731 & 0.822 \\
    ConvNeXt-T       & 32.34  & 0.924 & 0.866 & 0.973 & 0.829 & 0.446 & 0.749 & 0.837 \\
    ConvNeXt-B       & 92.35  & 0.928 & 0.871 & 0.974 & 0.846 & 0.462 & 0.760 & 0.845 \\
    Swin-T           & 32.04  & 0.929 & 0.873 & 0.974 & 0.847 & 0.469 & 0.763 & 0.848 \\
    Swin-B           & 91.53  & 0.925 & 0.866 & 0.973 & 0.828 & 0.454 & 0.752 & 0.839 \\
    \bottomrule
  \end{tabular}
  }
  }
\end{table*}

%% file: tables/paired_multimodal.tex
\begin{table*}[t]
  \centering
  \caption{Results of scenarios on \datasetns: the single-image baselines (i); paired double-pass (ii), and fusion-based (iii).}
  \label{tab:paired_multimodal}
  {\scriptsize
  \setlength{\tabcolsep}{4pt}
  \renewcommand{\arraystretch}{1.05}
  \setlength{\aboverulesep}{0pt}
  \setlength{\belowrulesep}{0pt}
  \begin{tabular}{@{}lcc|cc|ccc|cc@{}}
    \toprule
    & & & \multicolumn{2}{c|}{\textbf{Binary}} & \multicolumn{5}{c}{\textbf{Multiclass}} \\
    \textbf{Scenarios} & \textbf{Enc.} & \textbf{Input} & \textbf{F1} & \textbf{IoU} & \textbf{IoU}$_\mathrm{bg}$ & \textbf{IoU}$_\mathrm{perm}$ & \textbf{IoU}$_\mathrm{flood}$ & \textbf{IoU}$_\mathrm{avg}$ & \textbf{F1}$_\mathrm{avg}$ \\
    \midrule
    \midrule
    \multicolumn{10}{l}{\textit{(i) Single-image}} \\
    \midrule
    Post-hoc 3-class & Fr. & S1 & 0.931 & 0.877 & 0.975 & 0.849 & 0.478 & 0.767 & 0.851 \\
    Post-hoc 3-class & FT & S1 & 0.932 & 0.878 & 0.975 & 0.856 & 0.479 & 0.771 & 0.853 \\
    \midrule
    \midrule
    \multicolumn{10}{l}{\textit{(ii) Paired, two-pass}} \\
    \midrule
    Flood loss & Fr. & S1               & 0.929 & 0.873 & 0.974 & 0.840 & 0.465 & 0.760 & 0.845 \\
    Flood loss & Fr. & S2$\rightarrow$S1 & 0.920 & 0.858 & 0.972 & 0.839 & 0.430 & 0.747 & 0.833 \\
    Flood loss & Fr. & S1+S2$\rightarrow$S1 & 0.924 & 0.865 & 0.973 & 0.854 & 0.449 & 0.759 & 0.843 \\
    Flood loss & FT  & S1               & 0.933 & 0.879 & 0.975 & 0.860 & 0.486 & 0.774 & 0.855 \\
    Flood loss & FT & S2$\rightarrow$S1 & 0.931 & 0.876 & 0.975 & 0.868 & 0.486 & 0.776 & 0.857 \\
    Flood loss & FT & S1+S2$\rightarrow$S1 & 0.929 & 0.872 & 0.974 & 0.863 & 0.491 & 0.776 & 0.857 \\
    \midrule
    \midrule
    \multicolumn{10}{l}{\textit{(iii) Paired, single-pass (fusion)}} \\
    \midrule
    Early fusion & Fr. & S1    & 0.921 & 0.861 & 0.972 & 0.844 & 0.407 & 0.741 & 0.827 \\
    Mid fusion   & Fr. & S1    & 0.925 & 0.867 & 0.973 & 0.833 & 0.476 & 0.761 & 0.847 \\
    Early fusion & Fr. & S2$\rightarrow$S1 & 0.933 & 0.878 & 0.975 & 0.897 & 0.443 & 0.772 & 0.849 \\
    Mid fusion & Fr. & S2$\rightarrow$S1 & 0.936 & 0.883 & 0.976 & 0.895 & 0.467 & 0.779 & 0.856 \\
    Early fusion & Fr. & S1+S2$\rightarrow$S1 & 0.929 & 0.873 & 0.974 & 0.898 & 0.440 & 0.771 & 0.848 \\
    Mid fusion   & Fr. & S1+S2$\rightarrow$S1 & 0.937 & 0.887 & 0.977 & 0.896 & 0.488 & 0.787 & 0.863 \\
    Early fusion & FT  & S1    & 0.937 & 0.886 & 0.977 & 0.872 & 0.494 & 0.781 & 0.861 \\
    Mid fusion   & FT  & S1    & 0.937 & 0.886 & 0.977 & 0.885 & 0.490 & 0.784 & 0.862 \\
    Early fusion & FT  & S2$\rightarrow$S1 & \textbf{0.942} & \textbf{0.895} & \textbf{0.979} & \underline{0.906} & \textbf{0.521} & \textbf{0.802} & \textbf{0.875} \\
    Mid fusion   & FT & S2$\rightarrow$S1 & 0.940 & 0.890 & \underline{0.978} & 0.905 & 0.499 & 0.794 & 0.868 \\
    Early fusion & FT  & S1+S2$\rightarrow$S1 & \underline{0.941} & 0.892& \underline{0.978} & 0.902 & \textbf{0.521} & 0.800 & \underline{0.874} \\
    Mid fusion   & FT  & S1+S2$\rightarrow$S1 & \textbf{0.942} & \underline{0.894} & \textbf{0.979} & \textbf{0.911} & \underline{0.513} & \underline{0.801} & 0.873 \\
    \bottomrule
  \end{tabular}
  }
\end{table*}

%% file: tables/modality_ablation.tex
\begin{table}[t]
  \centering
  \caption{Modality ablation results on \datasetns with different input combinations.}
  \label{tab:modality_ablation}
  \scriptsize
  \setlength{\tabcolsep}{4pt}
  \renewcommand{\arraystretch}{1.05}
  \setlength{\aboverulesep}{0pt}
  \setlength{\belowrulesep}{0pt}
  \begin{tabular}{@{}ccc|cc@{}}
    \toprule
    \textbf{SAR} & \textbf{S2-L2A} & \textbf{DEM} & \textbf{F1$_{\mathrm{bin}}$} & \textbf{IoU$_{\mathrm{bin}}$} \\
    \midrule
    \midrule
    \multirow{4}{*}{S1 GRD} &   --     &    --    & $0.963 \scriptstyle \pm 0.002$ & $0.931 \scriptstyle \pm 0.003$ \\
                            & \cmark &    --    & $\textbf{0.972} \scriptstyle   \pm 0.001$ & $\underline{0.946} \scriptstyle \pm 0.001$ \\
                            &    --    & \cmark & $0.965 \scriptstyle \pm 0.002$ & $0.934 \scriptstyle \pm 0.003$ \\
                            & \cmark & \cmark & $\textbf{0.972} \scriptstyle \pm 0.001$ & $\textbf{0.947} \scriptstyle \pm 0.002$ \\
    \midrule
    \multirow{4}{*}{S1 RTC} &    --    &    --    & $0.958 \scriptstyle \pm 0.002$ & $0.922 \scriptstyle \pm 0.004$ \\
                            & \cmark &    --    & $0.965 \scriptstyle \pm 0.001$    &  $0.935 \scriptstyle \pm 0.003$ \\
                            &   --     & \cmark & $0.960 \scriptstyle \pm 0.004$ & $0.924 \scriptstyle \pm 0.006$ \\
                            & \cmark & \cmark & $0.970 \scriptstyle \pm 0.001$ & $0.942 \scriptstyle \pm 0.004$ \\
    \bottomrule
  \end{tabular}
\end{table}

%% file: tables/cross_dataset_comparison.tex
\begin{table*}[t]
  \centering
  \caption{Cross-dataset generalization results, training on the selected dataset and testing on the \dataset held-out set, composed of flood events in 2026.}
  \label{tab:cross_dataset_comparison}
  {\scriptsize
  \setlength{\tabcolsep}{4pt}
  \renewcommand{\arraystretch}{1.05}
  \setlength{\aboverulesep}{0pt}
  \setlength{\belowrulesep}{0pt}
  \begin{tabular*}{\textwidth}{@{\extracolsep{\fill}}l|cc|ccc|cc@{}}
    \toprule
    & \multicolumn{2}{c|}{\textbf{Binary}} & \multicolumn{5}{c}{\textbf{Multiclass}} \\
    \textbf{Training set} & \textbf{F1} & \textbf{IoU} & \textbf{IoU}$_\mathrm{bg}$ & \textbf{IoU}$_\mathrm{perm}$ & \textbf{IoU}$_\mathrm{flood}$ & \textbf{IoU}$_\mathrm{avg}$ & \textbf{F1}$_\mathrm{avg}$ \\
    \midrule
    \midrule
    \multicolumn{8}{l}{\textit{Frozen encoder}} \\
    \midrule
    MMFlood~\cite{montello2022mmflood}                 & 0.873 & 0.790 & 0.954 & 0.520 & 0.512 & 0.662 & 0.779 \\
    Sen1Floods11~\cite{bonafilia2020sen1floods11}      & 0.851 & 0.759 & 0.938 & 0.499 & 0.463 & 0.633 & 0.755 \\
    WorldFloods v2~\cite{portalesjulia2023worldfloods} & 0.882 & 0.802 & 0.954 & 0.655 & 0.515 & 0.708 & 0.816 \\
    Kuro Siwo~\cite{bountos2023kurosiwo}                & 0.887 & 0.809 & 0.963 & 0.543 & 0.568 & 0.691 & 0.803 \\
    \textbf{\dataset} & \underline{0.911} & \underline{0.845} & \underline{0.971} & \underline{0.709} & \underline{0.590} & \underline{0.757} & \underline{0.852} \\
    \midrule
    \midrule
    \multicolumn{8}{l}{\textit{Finetuned encoder}} \\
    \midrule
    MMFlood~\cite{montello2022mmflood}                 & 0.758 & 0.659 & 0.943 & 0.202 & 0.380 & 0.509 & 0.619 \\
    Sen1Floods11~\cite{bonafilia2020sen1floods11}      & 0.879 & 0.797 & 0.956 & 0.648 & 0.504 & 0.703 & 0.811 \\
    WorldFloods v2~\cite{portalesjulia2023worldfloods} & 0.877 & 0.795 & 0.954 & 0.650 & 0.492 & 0.698 & 0.808 \\
    Kuro Siwo~\cite{bountos2023kurosiwo}                & 0.888 & 0.811 & 0.957 & 0.635 & 0.544 & 0.712 & 0.820 \\
    \textbf{\dataset}                           & \textbf{0.917} & \textbf{0.854} & \textbf{0.972} & \textbf{0.716} & \textbf{0.601} & \textbf{0.763} & \textbf{0.857} \\
    \bottomrule
  \end{tabular*}
  }
\end{table*}

%% file: sections/6_concl.tex
\section{Conclusion}
\label{sec:concl}

We introduced \datasetns, a large-scale multi-modal flood benchmark from Copernicus EMS Rapid
Mapping activations.
Benchmarking geospatial foundation models against ImageNet-pretrained encoders, we find that training design matters more than the encoder: foundation models hold only a modest edge, temporal pairing alone does not help, and the best results come from end-to-end change-focused
architectures and optical-SAR fusion.
Training on \dataset also transfers to unseen events better than every benchmark we evaluate, in both regimes.
Some limitations remain: coverage is geographically skewed towards Europe (140 of 219 events), labels inherit residual noise from CEMS delineations and the DL-derived permanent-water layer, and flooded water, the rarest class, stays the hardest throughout. The cross-dataset comparison is moreover scored against \dataset's labels, so we anchor its claim on binary water delineation, where source conventions converge. Within this scope, optical context helps only before the event and the RTC and DEM layers add no measurable gain with the encoders tested, though purpose-built architectures may yet exploit them. By releasing \dataset with all modalities and a dedicated permanent-water layer, we provide a benchmark on which these gaps can be addressed.

%% file: sections/X_suppl.tex
\section{Models and Optimization}
\label{sec:suppl:implementation}

This appendix details how the models benchmarked in \cref{sec:method} are built,
trained, and scored. All runs are implemented in
TerraTorch~v1.1~\cite{gomes2025terratorch} with PyTorch Lightning in
\texttt{bf16-mixed} precision under a fixed global seed. They share the optimizer,
schedule, loss, augmentation, and tiled-inference protocol described below, and
differ only in the encoder, the decoder family, the learning-rate regime, and the
scenario-specific task head.

\subsection{Encoder zoo}
\label{sec:suppl:zoo}
We evaluate the geospatial foundation models TerraMind~v1 (tiny, small, base,
large)~\cite{jakubik2025terramind}, DOFA (base, large)~\cite{xiong2024dofa},
OlmoEarth~\cite{herzog2025olmoearth}, SSL4EO-ResNet50~\cite{wang2023ssl4eos12},
and Satlas Swin-B~\cite{bastani2023satlaspretrain}, alongside the
ImageNet-pretrained encoders ResNet-50/101~\cite{he2016resnet},
ConvNeXt-Tiny/Base~\cite{liu2022convnext}, and
Swin-Tiny/Base~\cite{liu2021swin}.
Each geospatial foundation model is run in two settings: with \emph{frozen
features}, where only the decoder and head are updated, and \emph{finetuned},
where the encoder and decoder are updated jointly. The ImageNet-pretrained
encoders are always trained end-to-end, with no frozen components.

\subsection{Decoder and segmentation head}
\label{sec:suppl:decoder}
Each foundation-model encoder is coupled with a U-Net~\cite{ronneberger2015unet}
decoder (channel widths $[512, 256, 128, 64]$) and a segmentation head with
dropout $0.3$.
For the transformer foundation models, the token outputs are converted into the
spatial feature pyramid the decoder expects through backbone-specific necks:
four intermediate blocks are selected, their token sequences are reshaped to
2-D feature maps, and a learned interpolation produces a four-level pyramid; the convolutional SSL4EO-ResNet50 exposes its native four stages directly. The Swin baseline needs only a dimension-permutation neck and $224{\times}224$ inputs.
The head emits per-pixel logits over $2$ classes for the binary
scenarios~(i)--(ii) and $3$ classes (background, permanent water, flooded water)
for the change-focused fusion architectures~(iii).

\subsection{Optimization}
\label{sec:suppl:optim}
All models are trained for $20$ epochs with AdamW~\cite{loshchilov2019adamw} under
a cosine-annealed learning rate ($\eta_{\min}=10^{-6}$). We use three
learning-rate regimes matched to the training mode:
\begin{itemize}
  \item \textbf{Frozen foundation models} (decoder and head only): learning rate
        $5\times10^{-6}$, weight decay $0.1$.
  \item \textbf{Finetuned foundation models}: decoder learning rate
        $5\times10^{-5}$ with a discriminative, $10\times$ lower encoder rate of
        $5\times10^{-6}$, weight decay $0.1$.
  \item \textbf{ImageNet-pretrained encoders} (end-to-end): decoder learning rate
        $5\times10^{-4}$ with an encoder rate of $5\times10^{-5}$, weight decay
        $0.01$.
\end{itemize}
The objective is pixel-wise cross-entropy with index $255$ ignored, so that pixels outside the Copernicus EMS analyzed area and invalid pixels (e.g.\ SAR no-data) are excluded from the loss. We train on $256{\times}256$ crops at stride $128$ (training tiles only) with D4 geometric augmentation, using a batch size of 8–64 depending on encoder and modality count. Training uses early stopping with patience 5 on the validation IoU of the positive class (binary water for scenarios (i)–(ii), flooded water for the three-class fusion models); the reported checkpoint is the one maximizing that same validation IoU, binary water IoU for the single-image and paired models, flooded-water IoU for the fusion models.
Inputs are normalized per channel: Sentinel-1 GRD VV/VH use our training-set
statistics ($\mu_{\mathrm{VV}}=-12.6$, $\sigma_{\mathrm{VV}}=5.2$;
$\mu_{\mathrm{VH}}=-20.3$, $\sigma_{\mathrm{VH}}=5.9$, in dB), except backbones that ship their own published SAR statistics (e.g.\ SSL4EO-ResNet50), which use those.

\subsection{Inference and tiling}
\label{sec:suppl:inference}
At test time, each $1024{\times}1024$ test tile is partitioned into a $4{\times}4$ grid of non-overlapping $256{\times}256$ windows, and each window is scored in a single forward pass; metrics are accumulated over all windows. 
Models that predict three classes directly (scenario~(iii)) are scored on that output; for the single-image and paired binary models the three-class flood map is assembled post-hoc by combining the binary pre- and post-event water masks, as described in \cref{sec:results:eval}.

\section{Permanent Water Layer Generation}
\label{sec:permwater}

Accurately distinguishing pre-existing, permanent water bodies from transient
flood inundation is a prerequisite for flood delineation.
Rather than relying on an external product such as the JRC Global Surface
Water~\cite{pekel2016high}, we generate a per-scene permanent water mask
directly from annual geospatial embeddings, without requiring cloud-free
Sentinel-2 imagery at inference time.

The core motivation is that annual embeddings are derived from multi-temporal
composites spanning a full year, making them insensitive to the specific
imaging conditions of any single acquisition.
Permanent water bodies leave a stable imprint in these composites that is
qualitatively different from the transient signal of a flood event occurring
in the same year.
A lightweight model trained to decode this imprint should therefore produce a
permanent water prior that is both temporally robust and spatially precise,
and that can be applied on demand to any geographic extent covered by the
embedding catalogue.

\subsection{Training data}
\label{sec:permwater:data}
We train and evaluate on the \emph{Earth Surface Water} (ESW) dataset~\cite{luo2021esw}, which provides binary water/non-water labels for 95 globally distributed Sentinel-2 Level-2A scenes acquired in 2019.
Following the original split, we tile each scene into non-overlapping $256\times256$ pixel patches, yielding 788 training tiles and 307 test tiles spanning diverse geographic and climatic conditions.

Based on the spatial and temporal extents of the ESW dataset, we download the same areas of interest from two publicly available annual embedding sources: (i) AlphaEarth Foundations (AEF)~\cite{alphaearth}, 64-dimensional embeddings at \SI{10}{\metre} resolution, available globally from 2017 to 2025, and (ii) TESSERA~\cite{tessera}, 128-dimensional embeddings at \SI{10}{\metre} resolution, also available globally in a variable range, around 2017 to 2025.
Both sources are fetched for the calendar year matching the Sentinel-2
acquisition (2019 for the ESW dataset).

A key property of both embedding catalogues is that each annual embedding
aggregates multi-temporal observations from the entire year into a single
compact representation.
Permanent water bodies (e.g., rivers, lakes, reservoirs, coastal lagoons) produce
a distinctive and stable pattern in this annual composite that is markedly
different from ephemeral flood signals, seasonal moisture variation, or
cloud-shadow artifacts.

\subsection{Methodology}
\label{sec:permwater:method}
Since the input is already a spatially dense, semantically rich volume, we do not employ additional heavyweight encoders.
Instead, we design two simple lightweight decoders that operate directly on the embedding tensor.
The first is a linear probe, a $1\times1$ convolution mapping directly to logits, comprising only 65 parameters for AEF and 129 parameters for TESSERA; it tests how much information is linearly accessible in the raw embedding without any spatial aggregation.
The second is a shallow convolutional decoder ($1\times1 \rightarrow 3\times3 \rightarrow 1\times1$) with non-linearities, totalling 41\,281 parameters for AEF and 45\,377 parameters for TESSERA, whose spatial convolution aggregates information over a $5\times5$ effective receptive field and allows the model to sharpen predictions along water boundaries.

We evaluate two reference baselines that require no embedding features.
First, we include JRC-GSW~\cite{pekel2016high}, a training-free static product derived from the full 1984--2021 Landsat archive providing per-pixel water occurrence statistics at \SI{30}{\metre} resolution.
We binarise the occurrence layer at $\geq$75\%, a threshold commonly adopted in the literature for permanent water, and query it directly from the Microsoft Planetary Computer STAC catalogue.
Second, we train a DeepLabV3+~\cite{chen2018deeplabv3} model (ResNet-50 backbone, ${\approx}25$\,M parameters) on the six Sentinel-2 bands augmented with two NDWI variants and NDVI, following the setup of~\cite{torchgeo_esw}.
This provides a spectral segmentation reference trained with comparable computational resources; unlike the embedding models, it has full access to the reflectance signal of each input scene.

\subsection{Results}
\label{sec:permwater:results}
\textbf{Implementation details.} We tile each scene into non-overlapping $256\times256$ pixel patches following the original ESW split, comprising 788 training tiles and 307 test tiles.
All models are trained for 50 epochs with AdamW~\cite{loshchilov2019adamw}
($\eta=10^{-4}$, weight decay $10^{-4}$) and a cosine-annealing schedule.
\Cref{tab:permwater} reports test-set performance on the ESW dataset.

\textbf{Linear probes are already strong.}
Even without any spatial aggregation, the linear probes reach 0.883 F1 for AEF and 0.902 F1 for TESSERA.
That a single $1\times1$ convolution with fewer than 130 parameters attains this level confirms that both embedding spaces encode permanent water as a nearly linearly separable signal, a direct consequence of the annual temporal aggregation described above.

Adding the spatial decoder yields $+8.0$ F1 points for AEF (from 0.883 to 0.963) but only $+0.9$ F1 points for TESSERA (from 0.902 to 0.911).
AEF's more compact 64-dimensional space benefits from explicit neighbourhood aggregation to resolve boundary ambiguities, whereas TESSERA's 128-dimensional representation already encodes sufficient per-pixel context.

Notably, JRC-GSW achieves F1~=~0.874 without any training on the ESW dataset, already ahead of DeepLabV3+ and only 8.9 points below our best model.
This confirms that permanent water is an exceptionally stable signal: decades of Landsat observations accumulate into a reliable occurrence prior that generalizes well across scenes.
A good portion of the residual gap to AEF-MLP could also be attributed to \emph{resolution}: JRC-GSW operates at \SI{30}{\metre} while our embeddings produce predictions at \SI{10}{\metre}, enabling finer delineation of narrow rivers, canals, and coastal features that are missed or blurred at coarser scales.

DeepLabV3+ reaches only 0.773 F1, sitting below even the training-free JRC-GSW.
We do not claim this is a tight spectral upper bound: a more elaborate setup (larger backbone, heavy augmentation, or scene-specific finetuning) could improve performance, but this comparison reflects a realistic, resource-comparable regime. The key limitation of spectral models is rather generalizability: a model trained on a limited number of scenes might not transfer well to globally distributed AoIs, each acquired under different atmospheric, sensor, and surface conditions. Annual geospatial embeddings sidestep this thanks to their full-year multi-temporal composition, and a single lightweight model can be applied globally without retraining with comparable robustness.

\begin{table}[t]
  \centering
  \caption{%
    Water body delineation on the Earth Surface Water test set.
    The upper block provides baselines, including JRC-GSW as state-of-the-art reference water.
    The lower block provides embedding-based results.
    Best results in \textbf{bold}.
  }
  \label{tab:permwater}
  \begin{tabular}{llrcccc}
    \toprule
    \textbf{Encoder} & \textbf{Decoder} & \textbf{\#Params} & \textbf{F1} & \textbf{IoU} & \textbf{Precision} & \textbf{Recall} \\
    \midrule
    \multicolumn{2}{l}{JRC-GSW~\cite{pekel2016high} (occ.$\geq$75\%)} & -- & 0.874 & 0.776 & 0.900 & 0.849 \\
    ResNet50 & DeepLabV3+ & 25.6M & 0.773 & 0.692 & 0.837 & 0.794 \\
    \midrule
    AEF     & Linear &    65 & 0.883 & 0.791 & \underline{0.906} & 0.862 \\
    TESSERA & Linear &   129 & 0.902 & 0.821 & 0.844 & \underline{0.969} \\
    TESSERA & MLP    &   45K & \underline{0.911} & \underline{0.837} & 0.852 & \textbf{0.980} \\
    AEF     & MLP    &   41K & \textbf{0.963} & \textbf{0.928} & \textbf{0.956} & \underline{0.969} \\
    \bottomrule
  \end{tabular}
\end{table}

\begin{figure*}[t]
  \centering
  \includegraphics[width=\textwidth]{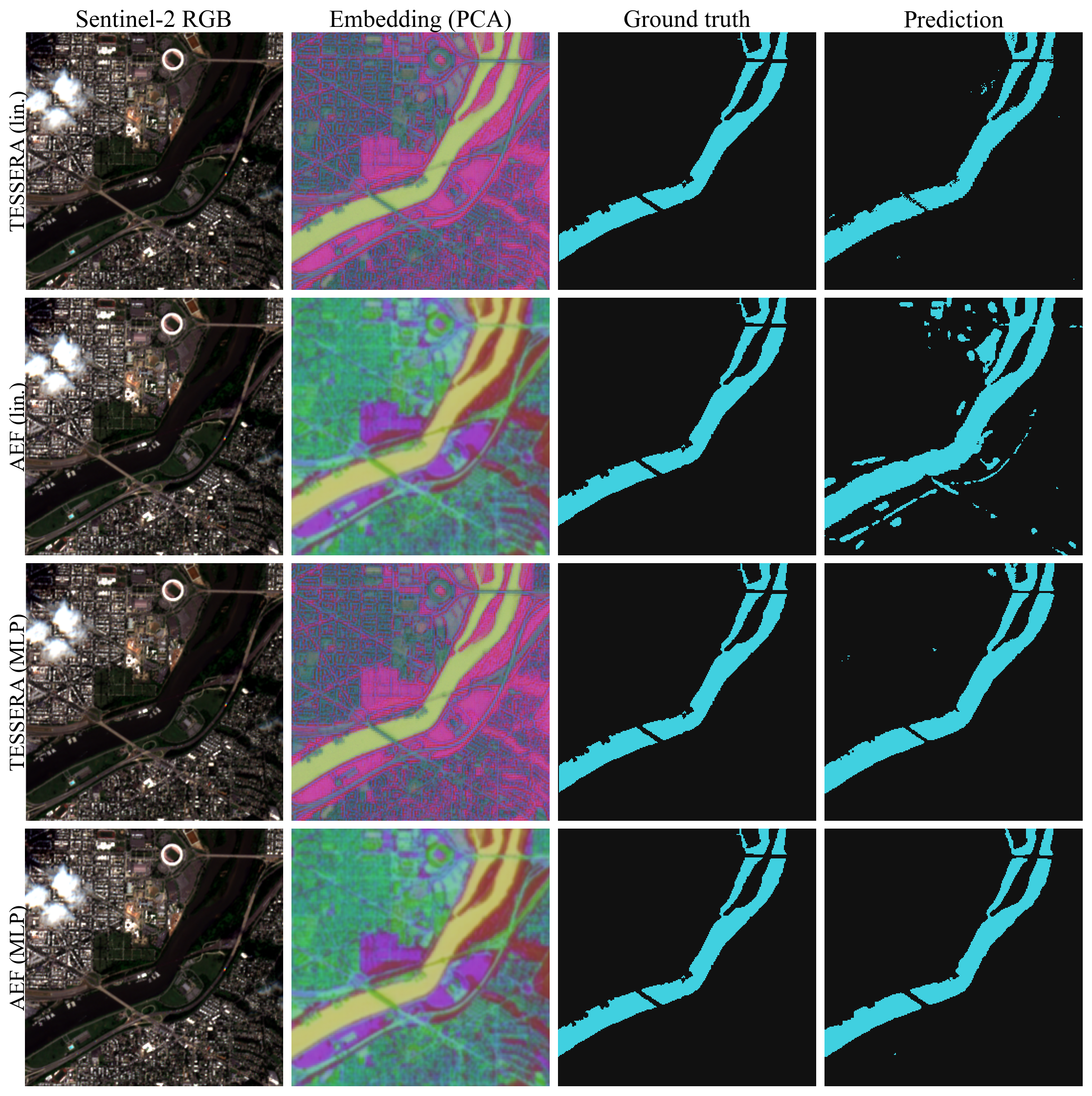}
  \caption{%
    Qualitative permanent water delineation on an ESW test tile.
    Rows correspond to the four embedding models (TESSERA and AEF, each with a
    linear probe and an MLP decoder); columns show the Sentinel-2 RGB
    composite, a PCA projection of the annual embedding, the ground-truth water
    mask, and the model prediction.
  }
  \label{fig:permwater_examples}
\end{figure*}

Qualitative examples are shown in \cref{fig:permwater_examples}.
The AEF-MLP model correctly delineates rivers, narrow channels, and coastal
features despite never observing any spectral band of the input scene.

\subsection{Inference Pipeline}
\label{sec:permwater:inference}
For each area of interest (AoI) in the flood dataset, we fetch the AEF
embedding corresponding to the year of the flood event and run the AEF-MLP
model to produce a binary permanent water mask at \SI{10}{\metre} resolution.
Events predating the AEF coverage window (before 2017) use the earliest
available year as a proxy; permanent water bodies are stable over multi-year
periods, so this introduces negligible error.
Binarisation uses hysteresis thresholding ($p \geq 0.5$ as seeds,
$p \geq 0.3$ for spatial extension) to prevent fragmentation of narrow rivers
and elongated reservoirs.

The pipeline was applied to all valid AoIs spanning the 219 flood events
of \cref{sec:dataset} (Copernicus EMS activations EMSR151--EMSR871), producing permanent water
layers at \SI{10}{\metre} resolution, three times finer than the JRC Global
Surface Water product (\SI{30}{\metre}) and without requiring access to a
multi-decadal Landsat archive.

